\documentclass[runningheads]{llncs}

\usepackage[numbers,sort&compress,sectionbib]{natbib}

\usepackage{microtype}

\usepackage{amsmath,amssymb,amsfonts}
\usepackage{bbm}

\usepackage{booktabs}

\usepackage{graphicx}
\usepackage{float}

\usepackage{paralist}

\usepackage{hyperref,cleveref}
\hypersetup{hidelinks}

\DeclareMathOperator*{\argmax}{arg\,max}

\usepackage{xcolor}

\begin{document}
\title{Neural Class Expression Synthesis\texorpdfstring{\thanks{This work is part of a project that has received funding from the European Union's Horizon 2020 research and innovation programme under the Marie Skłodowska-Curie grant No 860801 and the European Union’s Horizon Europe research and innovation programme under the grant No 101070305.
This work has also been supported by the Ministry of Culture and Science of North Rhine-Westphalia (MKW NRW) within the project SAIL under the grant No NW21-059D and by the Deutsche Forschungsgemeinschaft (DFG, German Research Foundation): TRR 318/1 2021~–~438445824.}}{}}
\author{N'Dah Jean Kouagou\orcidID{0000-0002-4217-897X} \and
Stefan Heindorf\orcidID{0000-0002-4525-6865} \and Caglar Demir\orcidID{0000-0001-8970-3850} \and Axel-Cyrille Ngonga Ngomo\orcidID{0000-0001-7112-3516}}
\authorrunning{N. J. Kouagou et al.}
\institute{Paderborn University, Germany\\
\email{\{ndah.jean.kouagou, heindorf, caglar.demir, axel.ngonga\}@upb.de}}

\maketitle
\begin{abstract}
Many applications require explainable node classification in knowledge graphs. Towards this end, a popular ``white-box'' approach is class expression learning: Given sets of positive and negative nodes, class expressions in description logics are learned that separate positive from negative nodes. Most existing approaches are search-based approaches generating many candidate class expressions and selecting the best one. However, they often take a long time to find suitable class expressions. In this paper, we cast class expression learning as a translation problem and propose a new family of class expression learning approaches which we dub neural class expression synthesizers. Training examples are ``translated'' into class expressions in a fashion akin to machine translation. Consequently, our synthesizers are not subject to the runtime limitations of search-based approaches. We study three instances of this novel family of approaches based on LSTMs, GRUs, and set transformers, respectively. An evaluation of our approach on four benchmark datasets suggests that it can effectively synthesize high-quality class expressions with respect to the input examples in approximately one second on average. Moreover, a comparison to state-of-the-art approaches suggests that we achieve better F-measures on large datasets. For reproducibility purposes, we provide our implementation as well as pretrained models in our public GitHub repository at \url{https://github.com/dice-group/NeuralClassExpressionSynthesis}
\keywords{Neural network \and Concept learning \and Class expression learning \and Learning from examples.}
\end{abstract}
\section{Introduction}
One of the most popular families of web-scale knowledge bases~\cite{heist2020knowledge} is that of RDF knowledge bases equipped with an ontology in W3C's web ontology language OWL~\cite{mcguinness2004owl}. Examples include DBpedia~\cite{bizer2009dbpedia}, Wikidata~\cite{vrandevcic2014wikidata}, and CaliGraph~\cite{heist2019uncovering}.
One means to implement ante-hoc explainable machine learning on these knowledge bases is class expression learning (also called concept learning)~\cite{fanizzi2008dl,lehmann2010concept,lehmann2010learning,rizzo2020class,tran2014bisimulation,heindorf2022evolearner}.
Informally, class expression learning approaches learn a class expression that describes individuals provided as positive examples. 
Class expression learning has 
applications in several domains, including ontology engineering 
\cite{lehmann2011class}, bio-medicine \cite{lehmann2014perspectives} and Industry 4.0 \cite{bin2017implementing}. There exist three main learning settings in class expression learning: 
\begin{inparaenum}[(1)]
\item positive and negative learning, 
\item positive-only learning, and 
\item class-inclusion learning~\cite{lehmann2009dl}.
\end{inparaenum} This paper tackles setting (1).

Several methods have been proposed to solve class expression learning problems; the best are based on refinement operators~\cite{fanizzi2008dl,lehmann2009dl,lehmann2010learning,lehmann2010concept,rizzo2020class,kouagou2022learning} and evolutionary algorithms~\cite{heindorf2022evolearner}. A common drawback of these approaches is their lack of scalability. While the reasoning complexity of all learning approaches grows with the expressivity of the underlying description logic (DL)~\cite{ozaki2020learning,konev2016model}, those based on refinement operators and evolutionary algorithms further suffer from the exploration of an infinite conceptual space for each learning problem~\cite{rizzo2020class}. 
Another inherent limitation of existing methods for class expression learning from examples is their inability to leverage previously solved problems\textemdash their algorithm always starts from scratch for each new learning problem.

In view of the large sizes of modern knowledge bases, e.g., DBpedia~\cite{bizer2009dbpedia} and Wikidata~\cite{vrandevcic2014wikidata}, we propose a new family of approaches, dubbed \textbf{neural class expression synthesizers} (NCES), for web-scale applications of class expression learning. The fundamental hypothesis behind this family of algorithms is that one should be able to capture enough semantics from latent representations (e.g., embeddings) of examples to directly synthesize class expressions in a fashion akin to machine translation, i.e., without the need for costly exploration. This hypothesis is supported by the significant improvement in the performance of machine translation approaches brought about by neural machine translation (NMT)~\cite{cho2014properties,wu2016google}. NMT approaches translate from a source language to a target language by exploiting an intermediary representation of a text's semantics. NCES behave similarly but translate from the ``language'' of sets of positive/negative examples to the ``language'' of class expressions. We instantiate this new paradigm by implementing three NCES instances that target the description logic $\mathcal{ALC}$. We show that our NCES instances generate high-quality class expressions with respect to the given sets of examples while remaining scalable. In fact, NCES instances synthesize solutions for multiple learning problems at the same time as they accept batches of inputs. This makes NCES particularly fit for deployment in large-scale applications of class expression learning, e.g., on the web.

The rest of the paper is organized as follows: First, we present existing approaches for class expression learning and introduce the notations and prerequisites needed throughout the paper. Next, we describe the intuition behind NCES in detail and introduce three instantiations of this new family of algorithms. We then compare these instantiations with state-of-the-art approaches on four benchmark datasets. Finally, we discuss our results and draw conclusions from our experiments.

\section{Related Work}
Class expression learning has been of interest to many researchers in recent years. 
Of the proposed approaches, the most prominent include those based on evolutionary algorithms~\cite{heindorf2022evolearner} and refinement operators~\cite{fanizzi2008dl,lehmann2010learning,lehmann2011class,buhmann2016dl,sarker2019efficient,kouagou2022learning}. The state-of-the-art EvoLearner~\cite{heindorf2022evolearner} initializes its population by random walks on the knowledge graph which are subsequently converted to description logic concepts, represented as abstract syntax trees. These concepts are further refined by means of mutation and crossover operations. EvoLearner outperformed approaches based on refinement operators such as CELOE and OCEL from the DL-Learner framework~\cite{heindorf2022evolearner}. Previously,~\citet{lehmann2010concept} studied different properties that a refinement operator can have, then designed a refinement operator to learn class expressions in description logics. Their learning algorithm, CELOE~\cite{lehmann2011class}, is implemented in DL-Learner~\cite{lehmann2009dl} alongside OCEL and ELTL~\cite{buhmann2016dl}. CELOE extends upon OCEL by using a different heuristic function and is currently regarded as the best class expression learning algorithm in DL-Learner. Although ELTL was designed for the lightweight description logic $\mathcal{EL}$, we include it in this study to check whether our generated learning problems can be solved in a simpler description logic. ECII~\cite{sarker2019efficient} is a recent approach for class expression learning that does not use a refinement operator and only invokes a reasoner once for each run. This approach was designed to overcome the runtime limitations of refinement operator-based approaches.
Other attempts to prune the search space of refinement operator-based approaches include
\textsc{DL-Focl}~\cite{rizzo2020class}. 
It is a modified version of \textsc{DL-Foil}~\cite{fanizzi2008dl} that is quintessentially based on omission rates.
\textsc{DL-Focl} uses techniques such as lookahead strategy and local memory to avoid reconsidering sub-optimal choices.

Even though existing approaches for class expression learning have shown promising results, most of them are search-based. As a result, these approaches often use entailment checks\textemdash which are hard to compute, see~\citet{ozaki2020learning}\textemdash or compute classification accuracies at each step of the search process. In contrast, NCES only require a refinement operator for generating training data but not at prediction time. Hence, NCES are particularly suitable for solving many different learning problems consisting of positive and negative examples on the same dataset. As the training process of our synthesizers only involves instance data embeddings and a vocabulary of atoms, they can be extended to more expressive description logics such as $\mathcal{ALCHIQ(D)}$.

\section{Background}
\subsection{Notation}
\label{notation}
 DL is short for description logic, and DNN stands for deep neural network. Unless otherwise specified, $\mathcal{K}=(\mathit{TBox}, \mathit{ABox})$ is a knowledge base in $\mathcal{ALC}$, and $\mathcal{N}_I$ is the set of all individuals in $\mathcal{K}$. The $\mathit{ABox}$ consists of statements of the form $C(a)$ and $R(a,b)$, whereas the $\mathit{TBox}$ contains statements of the form $C\sqsubseteq D$, where $C, D$ are concepts, $R$ is a role, and $a,b$ are individuals in $\mathcal{K}$. We use the representation of OWL knowledge bases as sets of triples to compute embeddings of individuals, classes and roles. The conversion into triples is carried out using standard libraries such as RDFLib~\cite{krech2006rdflib}. We then use the knowledge graph representation of the knowledge bases to compute embeddings, which are essential to our proposed approach (see Figure~\ref{fig:approach}). The function $|.|$ returns the cardinality of a set. $\mathbbm{1}$ denotes the indicator function, i.e., a function that takes two inputs and returns $1$ if they are equal, and $0$ otherwise. Let a matrix $M$ and integers $i$, $j$ be given. $M_{:,j}$, $M_{i,:}$, and $M_{ij}$ represent the $j$-th column, the $i$-th row, and the entry at the $i$-th row and $j$-th column, respectively. Similar notations are used for higher-dimensional tensors.
 
 We define the vocabulary $\mathit{Vocab}$ 
 of a given knowledge base $\mathcal{K}$ to be the list of all atomic concepts and roles in $\mathcal{K}$, together with the following constructs in any fixed ordering: `` '' (white space), ``.'' (dot), ``$\sqcup$'', ``$\sqcap$'', ``$\exists$'', ``$\forall$'', ``$\neg$'', ``('', and ``)'', which are all referred to as atoms. $\mathit{Vocab}[i]$ is the atom at position $i$ in $\mathit{Vocab}$. These constructs are used by NCES to synthesize class expressions in $\mathcal{ALC}$ (see Section~\ref{sec:nces} for details). Let $C$ be a class expression, then $\bar{C}$ and $\hat{C}$ are the list (in the order they appear in $C$) and set of atoms in $C$, respectively. 

\subsection{Description Logics}
Description logics~\cite{nardi2003introduction} are a family of knowledge representation paradigms based on first-order logics. They have applications in several domains, including artificial intelligence, the semantic web, and biomedical informatics.
In fact, the web ontology language, OWL, uses description logics to represent the terminological box of RDF ontologies.
In this work, we focus on the description logic $\mathcal{ALC}$ ($\mathcal{A}$ttributive $\mathcal{L}$anguage with $\mathcal{C}$omplement)~\cite{schmidt1991attributive} because of its simplicity and expressiveness. 
The syntax and semantics of $\mathcal{ALC}$ are presented in Table~\ref{tab:alc}.
\begin{table}[ht]
	\centering
	\caption{$\mathcal{ALC}$ syntax and semantics. $\mathcal{I}$ is an interpretation and $\Delta^\mathcal{I}$ its domain.}
	\setlength{\tabcolsep}{9.8pt}
  \begin{tabular}{@{}lcc@{}}
        \toprule
		\textbf{Construct} & \textbf{Syntax} & \textbf{Semantics}\\
		\midrule
        Atomic concept & $A$ & $A^{\mathcal{I}}\subseteq{\Delta^\mathcal{I}}$\\
        Atomic role & $R$ & 
		$R^\mathcal{I}\subseteq{\Delta^\mathcal{I}\times \Delta^\mathcal{I}}$\\
		Top concept & $\top$ & $\Delta^\mathcal{I}$\\
		Bottom concept & $\bot$& $\emptyset$\\
		Conjunction & $C\sqcap D$& $C^\mathcal{I}\cap D^\mathcal{I}$\\
		Disjunction & $C\sqcup D$& $C^\mathcal{I}\cup D^\mathcal{I}$\\
		Negation & $\neg C$ & $\Delta^\mathcal{I}\setminus C^\mathcal{I}$\\
		Existential role restriction & $\exists~ R.C$ & $\{a^\mathcal{I} \in \Delta^\mathcal{I}/\exists~ b^\mathcal{I} \in C^\mathcal{I}, (a^\mathcal{I},b^\mathcal{I})\in R^\mathcal{I}\}$\\
		Universal role restriction & $ \forall~ R.C$ & $\{a^\mathcal{I} \in \Delta^\mathcal{I}/\forall~ b^\mathcal{I}, (a^\mathcal{I},b^\mathcal{I})\in R^\mathcal{I}\Rightarrow b^\mathcal{I}\in C^\mathcal{I}\}$\\
		\bottomrule
	\end{tabular}
	\label{tab:alc}
\end{table}
\subsection{Refinement Operators}
\begin{definition}[\cite{lehmann2010concept}]
Given a quasi-ordered space $(\mathcal{S}, \preceq)$, a downward (respectively upward) refinement operator on $\mathcal{S}$ is a mapping $\rho: \mathcal{S}\rightarrow 2^\mathcal{S}$ such that for all $C\in \mathcal{S}$, $C' \in \rho(C)$ implies $C'\preceq C$ (respectively $C\preceq C'$).
\end{definition}
A refinement operator can be finite, proper, redundant, minimal, complete, weakly complete or ideal. Note that some of these properties can be combined whilst others cannot~\cite{lehmann2010concept}.
For class expression learning in description logics, weakly complete, finite, and proper refinement operators are the most used.

\subsection{Class Expression Learning}
\begin{definition}
\label{def:conceptlearning}
Given a knowledge base $\mathcal{K}$, a target concept $T$, a set of positive examples $E^+=\{e^+_1, e^+_2, \ldots, e^+_{n_1}\}$, and a set of negative examples $E^-=\{e^-_1, e^-_2, \ldots, e^-_{n_2}\}$, the learning problem is to find a class expression $C$ such that for $\mathcal{K}'=\mathcal{K} \cup \{T\equiv C\}$, we have $\forall$ $e^+ \in E^+\ \forall e^- \in E^-, \ \mathcal{K}' \models C(e^+)$ and $\mathcal{K}' \not\models C(e^-)$
. 
\end{definition}

Most existing approaches use hard-coded heuristics or refinement operators to search for the solution $C$. When an exact solution does not exist, an approximate solution in terms of accuracy or F-measure is to be returned by the approaches. In this work, we exploit the semantics embedded in latent representations of individuals (instance data) to directly synthesize $C$.

\subsection{Knowledge Graph Embedding}
A knowledge graph can be regarded as a collection of assertions in the form of subject-predicate-object triples $(s, p, o)$.
Embedding functions project knowledge graphs onto continuous vector spaces to facilitate downstream tasks such as link prediction~\cite{bordes2013translating}, recommender systems~\cite{zhang2016collaborative}, and structured machine learning~\cite{kouagou2022learning}. Many embedding approaches for knowledge graphs exist~\cite{wang2017knowledge,dai2020survey}. Some of them use only facts observed in the knowledge graph~\cite{nickel2012factorizing,bordes2014semantic}. Others leverage additional available information about entities and relations, such as textual descriptions~\cite{xie2016representation,wang2016text}. Most embedding approaches initialize each entity and relation with a random vector, matrix or tensor and learn the embeddings as an optimization problem. For example, TransE~\cite{bordes2013translating} represents entities and relations as vectors in the same space and aims to minimize the Euclidean distance between $s+p$ and~$o$ for each triple $(s, p, o)$. In this work, we use ConEx~\cite{demir2020convolutional} and TransE to evaluate NCES.

\subsection{Permutation-Invariant Network Architectures for Set Inputs}
We deal with set-structured input data as in 3D shape recognition~\cite{qi2017pointnet}, multiple instance learning~\cite{dietterich1997solving}, and few-shot learning~\cite{finn2017model,snell2017prototypical}. These tasks benefit from machine learning models that produce the same results for any arbitrary reordering of the elements in the input set. Another desirable property of these models is the ability to handle sets of arbitrary size.

In recent years, several approaches have been developed to meet the aforementioned requirements. The most prominent of these approaches include Deep Set~\citep{zaheer2017deep} and Set Transformer~\citep{lee2019set}. The Deep Set architecture encodes each element in the input set independently and uses a pooling layer, e.g., averaging, to produce the final representation of the set. In contrast, the Set Transformer architecture uses a self-attention mechanism to represent the set, which allows pair-wise and even higher-order interactions between the elements of the input set. As a result, the Set Transformer architecture shows superior performance on most tasks compared to Deep Set with a comparable model size~\cite{lee2019set}. In this work, we hence use the Set Transformer architecture (more details in Section~\ref{sec:nces}) and refer to \cite{lee2019set} for a description of its building blocks: Multi-head Attention Block (MAB), Set Attention Block (SAB), Induced Set Attention Block (ISAB), and Pooling by Multi-head Attention (PMA). 

\section{Neural Class Expression Synthesis}
\label{sec:nces}
In this section, we present our proposed family of approaches for class expression learning from examples. 
We start with a formal definition of the learning problem that we aim to solve, then present our proposed approach in detail.

\subsection{Learning Problem}
We adapt the classical definition of a
learning problem (see Definition~\ref{def:conceptlearning}) to our setting of class expression synthesis (Definition~\ref{def:conceptlearningnces}).
\begin{definition}
\label{def:conceptlearningnces}
Given a knowledge base $\mathcal{K}$, a set of positive examples $E^+=\{e^+_1, e^+_2, \ldots, e^+_{n_1}\}$, and a set of negative examples $E^-=\{e^-_1, e^-_2, \ldots, e^-_{n_2}\}$, the learning problem is to synthesize a class expression $C$ in $\mathcal{ALC}$ using atoms (classes and roles) in $\mathcal{K}$ that (ideally) accurately classifies the provided examples. 
\end{definition}
In theory, there can be multiple solutions to a learning problem under both Definition~\ref{def:conceptlearning} and Definition~\ref{def:conceptlearningnces}; our NCES generate only one. Moreover, the solution computed by a concept learner might be an approximation, e.g., there might be some false positives and false negatives. NCES aim to obtain high values for accuracy and F-measure.

\subsection{Learning Approach (NCES)}
\label{sec:learn-app}
\begin{figure*}
    \centering
    \includegraphics[width=\textwidth]{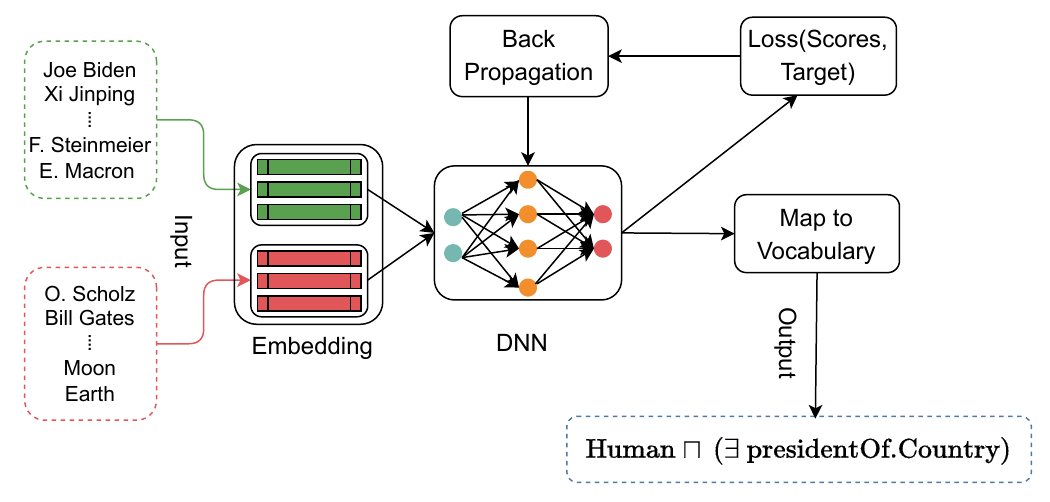}
    \caption{NCES architecture. DNN stands for deep neural network that produces a sequence of tokens in the vocabulary (e.g., a sequence-to-sequence or a set-to-sequence model). The input consists of positive examples (upper left, dotted green box) and negative examples (bottom left, dotted red box).}
    \label{fig:approach}
\end{figure*}

We propose the following recipe to implement the idea behind NCES. First, given a knowledge base over $\mathcal{ALC}$, convert it into a knowledge graph (see Subsection~\ref{notation}). Then, embed said knowledge graph into a continuous vector space using any state-of-the-art embedding model in the literature. In our experiments, we used two embedding models with different expressive power: ConEx which applies convolutions on complex-valued vectors, and TransE which projects entities and relations onto a Euclidean space and uses the Euclidean distance to model interactions. The computed embeddings are then used as features for a model able to take a set of embeddings as input and encode a sequence of atoms as output (see Figure~\ref{fig:approach}).
\subsubsection{Neural Network Architectures}
We conduct our experiments using the following network architectures: the Long Short-Term Memory (LSTM)~\citep{hochreiter1997long}, the Gated Recurrent Unit (GRU)~\citep{cho2014learning}, and the Set Transformer~\citep{lee2019set}. The latter is known to be permutation equivariant while the two others are not. Nonetheless, LSTM and GRU can handle set inputs as long as an ordering is defined since they deal well with sequential data~\citep{zhao2018sequential,zulqarnain2019efficient}. In this work, we use the default ordering (the order in which we received the data) of the elements in each set during the data generation process (see Section~\ref{sec:generate}).
\paragraph{Recurrent Networks (LSTM and GRU)}
We use two recurrent layers followed by three linear layers with the \texttt{relu} activation function and a batch normalization layer. A recurrent neural network produces a sequence of $n$ hidden states $h_i$ ($i$=$1,\dots,n$) for each input sequence of length $n$. In this work, we are concerned with a sequence of $n_1$ positive examples and a sequence of $n_2$ negative examples which are processed separately with the same network:
\begin{eqnarray}
h^{pos}_1, \dots, h^{pos}_{n_1} = \mathit{RNN}(x_{pos});\ h^{neg}_1, \dots, h^{neg}_{n_2} = \mathit{RNN}(x_{neg});
\end{eqnarray}
where $x_{pos}$ and $x_{neg}$ are the sequences of embeddings of positive and negative examples, respectively. $\mathit{RNN}$ is a two-layer LSTM or GRU network.
The hidden state vectors of the two sets of examples are summed separately, then concatenated and fed to a sequence of 3 linear layers:
\begin{align}
h_{pos} &:= \sum_{t=1}^{T_1} h^{pos}_t;\ h_{neg} := \sum_{t=1}^{T_2} h^{neg}_t;\ h := \mathrm{Concat}(h_{pos}, h_{neg});\\
\label{rnn_score}
O &= W_3(\mathrm{bn}(W_2f(W_1 h+b_1)+b_2)) + b_3.
\end{align}
Here, $f$ is the \texttt{relu} activation function, $\mathrm{bn}$ is a batch normalization layer, and $W_1, b_1, W_2, b_2, W_3, b_3$ are trainable weights. 

\paragraph{Set Transformer}
This architecture comprises an encoder $\mathit{Enc}$ and a decoder $\mathit{Dec}$, each with 4 attention heads. 
The encoder is a stack of two $\mathrm{ISAB}$ layers with $m=32$ inducing points, and the decoder is composed of a single $\mathrm{PMA}$ layer with one seed vector ($k=1$), and a linear layer. As in the previous paragraph, the sets of positive and negative examples for a given class expression are first encoded separately using the encoder. The outputs are then concatenated row-wise and fed to the decoder:
\begin{align}
O_{pos} &= \mathit{Enc}(x_{pos});\ O_{neg} = \mathit{Enc}(x_{neg}); \\
\label{st_score}
O &= \mathit{Dec}(\mathrm{Concat}(O_{pos}, O_{neg})).
\end{align}
Although the encoder captures interactions intra-positive and intra-negative examples separately, the decoder further captures interactions across the two sets of examples from the concatenated features through self-attention. This demonstrates the representational power of the Set Transformer model for our set-structured inputs for class expression synthesis.

The output $O$ from \ref{rnn_score} and \ref{st_score} is reshaped into a $(1+|\mathit{Vocab}|) \times L$ matrix, where $L$ is the length of the longest class expression that NCES instances can generate. These scores allow us to compute the loss (see Equation \ref{loss}) and update model weights through gradient descent during training.

\subsubsection{Loss}
We train our NCES instances using the loss function $\mathcal{L}$ defined by:
\begin{eqnarray}
\label{loss}
\mathcal{L}(x, y) = -\frac{1}{NL}\sum_{i=1}^N\sum_{j=1}^L \log\left(\frac{\exp(x_{i, y_{ij}, j})}{\sum_{c=1}^{C} \exp(x_{i, c, j})}\right),
\end{eqnarray}
where $N$ is the size of the minibatch, $C$ is the number of classes, $x \in \mathbb{R}^{N\times C \times L}$ is the minibatch of predicted class scores for each position in the target sequence of atoms, and $y \in \mathbb{N}^{N\times L}$ is the minibatch of actual class indices.
Minimizing $\mathcal{L}$ constrains the model to assign a high score to the entry corresponding to the correct token ($\exp(x_{i,y_{ij},j}) \approx 1$) while keeping the remaining scores relatively low ($\sum_{c=1, c\neq y_{ij}}^{C} \exp(x_{i, c, j}) \approx 0$).
In this work, $C=1+|\mathit{Vocab}|$, where the additional $+1$ accounts for the special token ``PAD'' that we used to pad all class expressions to the same length. Contrarily to some works that omit this special token when computing the loss, we use it as an ordinary token during training. This way, we can generate class expressions more efficiently at test time with a single forward pass in the model, then strip off the generated tokens after the special token. To avoid exploding gradients and accelerate convergence during training, we adopt the gradient clipping technique~\cite{zhang2019gradient}.

\subsubsection{Learning Metrics}
Apart from the loss function, we introduce two accuracy measures to quantify how well neural networks learn during training: soft accuracy and hard accuracy. The former only accounts for the correct selection of the atoms in the target expression, while the latter additionally measures the correct ordering of the selected atoms. Formally, let $T$ and $P$ be the target and predicted class expressions, respectively. Recall the notation $\bar{C}$ and $\hat{C}$ introduced in Section~\ref{notation} for any class expression $C$. The soft ($\mathit{Acc}_{s}$) accuracy and hard accuracy ($\mathit{Acc}_{h}$) are defined as follows:
\begin{eqnarray}
\mathit{Acc}_{s}(T, P) = \frac{|\hat{T}\cap \hat{P}|}{|\hat{T} \cup \hat{P}|};\
\mathit{Acc}_{h}(T, P) = \frac{\sum_{i=1}^{\min(l_1, l_2)} \mathbbm{1}(\bar{T}[i], \bar{P}[i])}{\max(l_1, l_2)}.
\end{eqnarray}
where $l_1$ and $l_2$ are the lengths of $\bar{T}$ and $\bar{P}$, respectively.

\subsubsection{Class Expression Synthesis}
\label{sec:select}
We synthesize class expressions by mapping the output scores $O$ (see Equations \ref{rnn_score} and \ref{st_score}) to the vocabulary. More specifically, we select the highest-scoring atoms in the vocabulary for each position along the sequence dimension:
\begin{align}
\label{synthesis1}
\mathrm{id}_j &= \argmax_{c\in \{1,\ldots, C\}}O_{c,j} \text{ for } j=1,\ldots, L,\\
\label{synthesis2}
\mathit{synthesized\_atom_j} &= \mathit{Vocab}[\mathrm{id}_j].
\end{align}

\subsubsection{Model Ensembling}
\label{sec:ensemble}
Ensemble learning has proven to be one of the most robust approaches for tasks involving complex noisy data~\cite{sagi2018ensemble,dong2020survey}. In this work, we combine class expression synthesizers' predictions post training by averaging the predicted scores. Specifically, given the output scores $O_i \in \mathbb{R}^{C\times L}$ ($i=1,2,3$) as defined in \ref{rnn_score} and \ref{st_score} for the three models LSTM, GRU, and Set Transformer, we consider four different ensemble models: three pairwise ensemble models, and one global ensemble model (LSTM, GRU, and Set Transformer are combined). Formally, the ensemble scores are computed as:
\begin{eqnarray}
O=\frac{\sum_{i\in\mathcal{I}}O_i}{|\mathcal{I}|} \text{ with }
\mathcal{I}\subseteq{\{1,2,3\}} \text{ and } |\mathcal{I}|\ge 2.
\end{eqnarray}
Then, the synthesized expression is constructed following Equations \ref{synthesis1} and \ref{synthesis2} using the average scores $O$.
\section{Evaluation}
\label{sec:eval}
\subsection{Experimental Setup}
\subsubsection{Datasets}
We evaluated our proposed approach on the Carcinogenesis~\cite{westphal2019sml}, Mutagenesis~\cite{westphal2019sml}, Semantic Bible\footnote{\url{https://www.semanticbible.com/ntn/ntn-overview.html}}, and the Vicodi~\cite{nagypal2005history} knowledge bases. Carcinogenesis and Mutagenesis are knowledge bases about chemical compounds and how they relate to each other. The Semantic Bible knowledge base describes each named object or thing in the New Testament, categorized according to its class, including God, groups of people, and locations. The Vicodi knowledge base was developed as part of a funded project and describes European history. The statistics of each of the knowledge bases are given in Table~\ref{tab:datasets}. 
\begin{table}[t]
\centering
\caption{Detailed information about the datasets used for evaluation. $\mathrm{|LPs|}$ is the number of learning problems in the test set.
}
 \setlength{\tabcolsep}{2.5pt}
  \begin{tabular}{@{}lcccccccc@{}}
    \toprule
    \textbf{Dataset} & $|$\textbf{Ind.}$|$ & $|$\textbf{Classes}$|$ & $|$\textbf{Prop.}$|$ & $|\textbf{TBox}|$ & $|\textbf{ABox}|$ & $|\textbf{Train}|$ & $|\textbf{LPs}|$ & $|\textbf{Vocab}|$\\
    \midrule
    Carcinogenesis & 22,372 & 142 &\phantom{0}4 & 144 &\phantom{0}74,223 & 10,982 & 111 & 157\\
    Mutagenesis & 14,145 &\phantom{0}86 &\phantom{0}5 & \phantom{0}82 &\phantom{0}47,722 & \phantom{0}5,333 & \phantom{0}54 & 102\\
    Semantic Bible &\phantom{000}724 &\phantom{0}48 & 29 &\phantom{0}56 &\phantom{00}3,106 & \phantom{0}3,896 & \phantom{0}40 &\phantom{0}88\\
    Vicodi & 33,238 & 194 & 10 & 204 & 116,181 & 18,243 & 175 & 215\\
    \bottomrule
  \end{tabular}
  \label{tab:datasets}
\end{table}

\subsubsection{Training and Test Data Construction}
\label{sec:generate}
We generated class expressions of different forms from the input knowledge base using the recent refinement operator by~\citet{kouagou2022learning} that was developed to efficiently generate numerous class expressions to serve as training data for concept length prediction in $\mathcal{ALC}$. The data that we generate is passed to the filtering process, which discards any class expression~$C$ such that an equivalent but shorter class expression~$D$ was not discarded. Note that each class expression comes with its set of instances, which are computed using the fast closed-world reasoner based on set operations described in~\cite{heindorf2022evolearner}. These instances are considered positive examples for the corresponding class expression; negative examples are the rest of the individuals in the knowledge base.
Next, the resulting data is randomly split into training and test sets; we used the discrete uniform distribution for this purpose. To ensure that our approach is scalable to large knowledge bases, we introduce a hyper-parameter $n=n_1+n_2$ that represents the total number of positive and negative examples we sample for each class expression to be learned by NCES. Note that $n$ is fixed for each knowledge base, and it depends on the total number of individuals.

\subsubsection{Evaluation Metrics}
We measure the quality of a predicted class expression in terms of accuracy and F-measure with respect to the positive/negative examples. Note that we cannot expect to exactly predict the target class expression in the test data since there can be multiple equivalent class expressions.

\subsubsection{Hyper-parameter Optimization for NCES}
We employed random search on the hyper-parameter space since it often yields good results while being computationally more efficient than grid search~\cite{bergstra2012random}; the selected values---those with the best results---are reported in Table~\ref{tab:hyperparam}. 
In the table, it can be seen that most knowledge bases share the same optimal values of hyper-parameters: the minibatch size $N$, the number of training epochs $epochs$, the optimizer $opt.$, the learning rate $lr$, the maximum output sequence length $L$, the number of embedding dimensions $d$, the number of inducing points $m$, and the gradient clipping value $gc$. Although we may increase $n$ for very large knowledge bases, $n=\min (\frac{|\mathcal{N}_I|}{2}, 1000)$ appears to work well with our evaluation datasets. This suggests that one can effortlessly find fitting hyper-parameters for new datasets.
\begin{table}[tb]
\centering
\caption{Hyper-parameter settings per dataset. Recall that $m$ is the number of inducing points in the Set Transformer model, and $n$ is the number of examples.}
\setlength{\tabcolsep}{7.6pt}
\begin{tabular}{@{}lccccccccc@{}}
    \toprule
    \textbf{Dataset} & \textbf{\textit{epochs}} & \textbf{\textit{opt.}} & \textbf{\textit{lr}} & \textbf{\textit{d}} & \textbf{\textit{N}} & \textbf{\textit{L}} & \textbf{\textit{n}} & \textbf{\textit{m}} & \textbf{\textit{gc}}\\
    \midrule
    Carcinogenesis & 300 & Adam & 0.001 & 40 & 256 & 48 & 1,000 & 32 & 5\\
    Mutagenesis & 300 & Adam & 0.001 & 40 & 256 & 48 & 1,000 & 32 & 5\\
    Semantic Bible & 300 & Adam & 0.001 & 40 & 256 & 48 &\phantom{00}362 & 32 & 5 \\
    Vicodi & 300 & Adam & 0.001 & 40 & 256 & 48 &1,000 & 32 & 5 \\
    \bottomrule
\end{tabular}
\label{tab:hyperparam}
\end{table}

\subsubsection{Hardware and Training Time}
We trained our chosen NCES instances on a server with 1TB of RAM and an NVIDIA RTX A5000 GPU with 24 GB of RAM. Note that during training, approximately 8GB of the 1TB RAM is currently used by NCES. As search-based approaches do not require a GPU for class expression learning, we used a 16-core Intel Xeon E5-2695 with 2.30GHz and 16GB RAM to run all approaches (including NCES post training) for class expression learning on the test set. The number of parameters and training time of each NCES instance are reported in Table~\ref{tab:runtime}. From the table, we can observe that NCES instances are lightweight and can be trained within a few hours on medium-size knowledge bases. Note that training is only required once per knowledge base.

\begin{table}[tb]
  \centering
  \caption{Model size and training time. The training time is in minutes.}
  \setlength{\tabcolsep}{3.5pt}
  \begin{tabular}{@{}lcccccccc@{}}
    \toprule
    &\multicolumn{2}{c}{\textbf{Carcinogenesis}} &\multicolumn{2}{c}{\textbf{Mutagenesis}} &\multicolumn{2}{c}{\textbf{Semantic Bible}}
    &\multicolumn{2}{c}{\textbf{Vicodi}}\\
    \cmidrule(lr){2-3}
    \cmidrule(lr){4-5}
    \cmidrule(lr){6-7}
    \cmidrule(lr){8-9}
    & $|\mathrm{Params.}|$ & $\mathrm{Time}$ & $|\mathrm{Params.}|$ & $\mathrm{Time}$ & $|\mathrm{Params.}|$ & $\mathrm{Time}$ & $|\mathrm{Params.}|$ & $\mathrm{Time}$\\
    \midrule
    NCES$_{LSTM}$ & 1,247,136 & 31.50 & 906,576 & 16.94 & 819,888 & \phantom{0}6.65 & 1,606,272 & 50.82\\
    NCES$_{GRU}$ & 1,192,352 & 21.61 & 851,792 & 12.28 & 765,104 & \phantom{0}5.39 & 1,551,488 & 34.15\\
    NCES$_{ST}$  & 1,283,104 & 40.82 & 942,544 & 21.36 & 855,856 & \phantom{0}7.98 & 1,642,240 & 66.19\\
    \bottomrule
  \end{tabular}
  \label{tab:runtime}
\end{table}

\subsection{Results}
\subsubsection{Syntactic Accuracy}
\begin{figure}[tbh]
\centering
  \includegraphics[width=\textwidth]{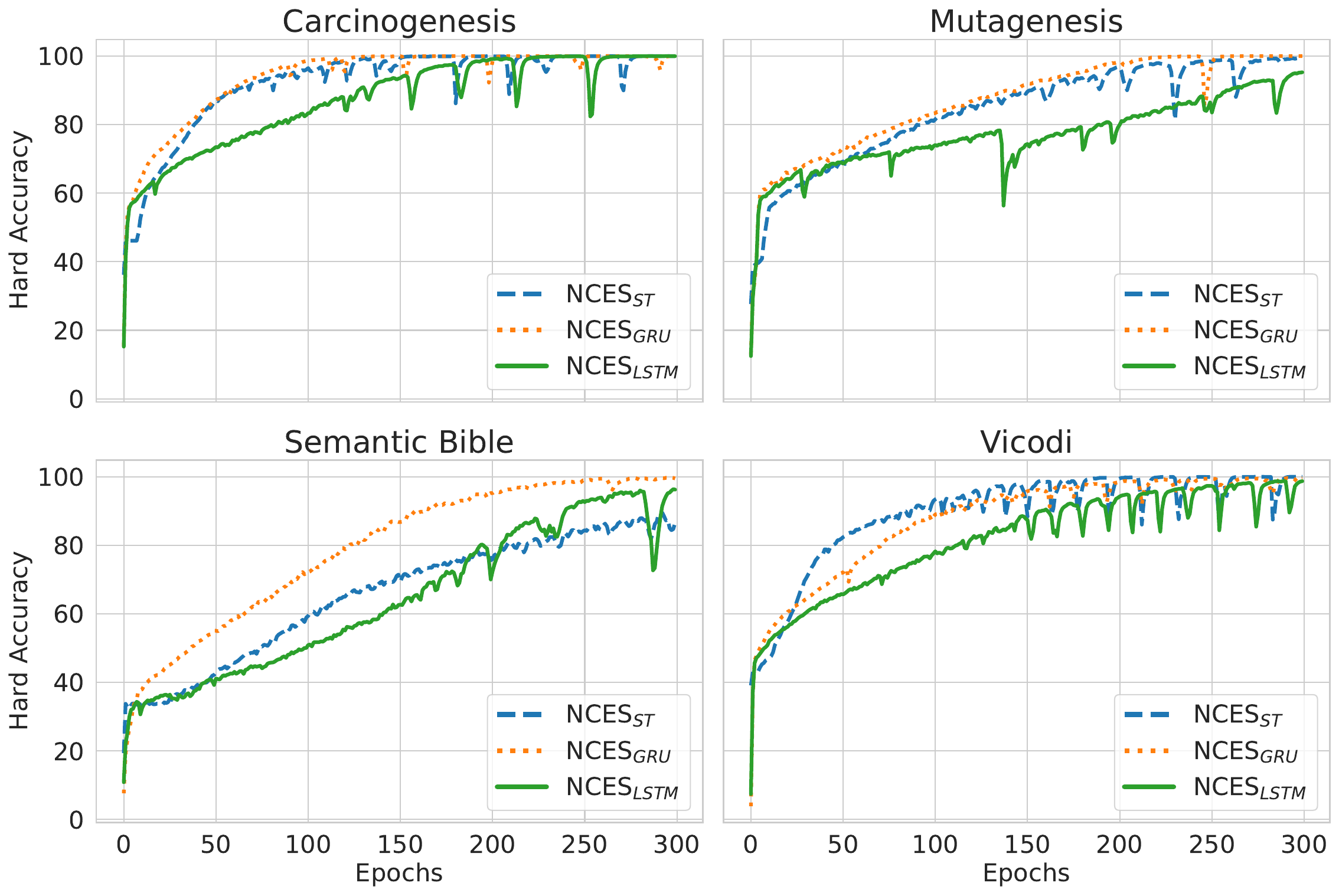}
\caption{Training accuracy curves.}
\label{fig:traing-curves}
\end{figure}

Our neural class expression synthesizers were trained for 300 epochs on each knowledge base. In Figure~\ref{fig:traing-curves}, we only show the hard accuracy curves during training due to space constraints. The rest of the training curves can be found on our GitHub repository. The curves in Figure~\ref{fig:traing-curves} suggest that NCES instances train fast with an exponential growth in accuracy within the first 10 epochs. All models achieve over 95\% syntactic accuracy on large knowledge bases (Carcinogenesis, Mutagenesis, and Vicodi). On the smallest knowledge base, Semantic Bible, we observe that NCES$_{ST}$ drops in performance as it only achieves 88\% accuracy during training. On the other side, NCES$_{GRU}$ and NCES$_{LSTM}$ tend to overfit the training data. This suggests that NCES instances are well suited for large datasets. We validate this hypothesis through the quality of the synthesized solutions on the test set (see Table~\ref{tab:evaluation-conex}).

\subsubsection{Comparison to State-of-the-art Approaches}
\begin{table}[tbp]
\centering
\caption{Evaluation results per approach and dataset. NCES uses ConEx embeddings. The star (*) indicates statistically significant differences between the best search-based and the best synthesis-based approaches. $\uparrow$ indicates that the higher is better, and $\downarrow$ indicates that the lower is better. Underlined values are the second best.}
\setlength{\tabcolsep}{1pt}
\footnotesize
  \begin{tabular}{@{}l@{\hskip 3pt}c@{\hskip 3pt}c@{\hskip 3pt}c@{\hskip 3pt}c@{}}
  \toprule
    & \multicolumn{4}{c}{\textbf{F$_1$ (\%) $\uparrow$}}\\
    \cmidrule(lr){2-5}
    & Carcinogenesis & Mutagenesis & Semantic Bible & Vicodi\\
    \midrule
    CELOE & 37.92$\pm$44.25 & 82.95$\pm$33.48 & \textbf{93.18$\pm$17.52}* & 35.66$\pm$42.06\\
    ELTL & 13.35$\pm$25.84 & 28.81$\pm$34.44 & 42.77$\pm$38.46 & 16.70$\pm$33.31\\
    ECII & 15.74$\pm$27.82 & 27.14$\pm$31.64 & 33.97$\pm$37.71 & 43.66$\pm$36.13\\
    EvoLearner & 91.48$\pm$14.30 & \textbf{93.27$\pm$12.95} & \underline{91.88$\pm$10.14} & \underline{92.74$\pm$10.28}\\
    \midrule
    NCES$_{LSTM}$ & 82.21$\pm$29.29 & 81.47$\pm$27.77 & 72.32$\pm$34.37 & 72.35$\pm$35.34\\
    NCES$_{GRU}$ & 89.51$\pm$25.24 & 78.24$\pm$30.99 & 52.37$\pm$39.86 & 86.60$\pm$26.84\\
    NCES$_{ST}$ & 90.32$\pm$25.12 & 80.55$\pm$34.33 & 72.95$\pm$38.68 & 77.75$\pm$37.16\\
    NCES$_{ST+GRU}$ & \underline{96.77$\pm$12.72} & 89.50$\pm$26.09 & 84.32$\pm$26.82 & 92.91$\pm$20.14\\
    NCES$_{ST+LSTM}$ & 96.72$\pm$12.96 & 89.19$\pm$26.16 & 80.78$\pm$28.30 & 90.91$\pm$21.57\\
    NCES$_{GRU+LSTM}$ & 94.51$\pm$15.39 & 81.49$\pm$30.71 & 75.91$\pm$32.48 & 88.28$\pm$24.17\\
    NCES$_{ST+GRU+LSTM}$ & \textbf{97.06$\pm$13.06}* & \underline{91.39$\pm$22.91} & 87.11$\pm$24.05 & \textbf{95.51$\pm$12.14}*\\
    \bottomrule
    \toprule
    & \multicolumn{4}{c}{\textbf{Accuracy (\%) $\uparrow$}}\\
    \cmidrule(lr){2-5}
    & Carcinogenesis & Mutagenesis & Semantic Bible & Vicodi\\
    \midrule
    CELOE & 66.88$\pm$24.87 & 94.39$\pm$11.68 & \underline{98.17$\pm$\phantom{0}4.95} & 85.38$\pm$17.71\\
    ELTL & 24.96$\pm$32.86 & 38.01$\pm$39.26 & 47.31$\pm$38.22 & 33.35$\pm$41.88\\
    ECII & 25.95$\pm$35.98 & 31.99$\pm$38.36 & 30.35$\pm$37.91 & 76.73$\pm$35.22\\
    EvoLearner & \underline{99.72$\pm$\phantom{0}1.44} & \textbf{99.35$\pm$2.03} & \textbf{98.89$\pm$3.45}* & \textbf{99.27$\pm$5.13}\\
    \midrule
    NCES$_{LSTM}$ & 97.92$\pm$11.19 & 98.04$\pm$\phantom{0}6.81 & 89.83$\pm$23.59 & 94.03$\pm$20.77\\
    NCES$_{GRU}$ & 99.32$\pm$\phantom{0}4.16 & 98.02$\pm$\phantom{0}4.03 & 80.67$\pm$30.80 & \underline{98.10$\pm$11.43}\\
    NCES$_{ST}$ & 97.16$\pm$14.33 & 90.71$\pm$25.49 & 84.19$\pm$32.72 & 89.90$\pm$27.89\\
    NCES$_{ST+GRU}$ & 99.03$\pm$\phantom{0}9.28 & 97.27$\pm$11.75 & 92.38$\pm$21.39 & 97.11$\pm$12.57\\
    NCES$_{ST+LSTM}$ & 99.26$\pm$\phantom{0}5.69 & 96.93$\pm$12.17 & 90.27$\pm$21.60 & 95.78$\pm$16.46\\
    NCES$_{GRU+LSTM}$ & \textbf{99.90$\pm$0.31} & \underline{98.49$\pm$\phantom{0}5.35} & 87.16$\pm$27.44 & 96.86$\pm$14.99\\
    NCES$_{ST+GRU+LSTM}$ & 99.04$\pm$\phantom{0}9.28 & 98.39$\pm$\phantom{0}6.45 & 94.70$\pm$17.83 & 97.48$\pm$11.71\\
    \bottomrule
    \toprule
    & \multicolumn{4}{c}{\textbf{Runtime (sec.) $\downarrow$}}\\
    \cmidrule(lr){2-5}
    & Carcinogenesis & Mutagenesis & Semantic Bible & Vicodi\\
    \midrule
    CELOE & 239.58$\pm$132.59 & 92.46$\pm$125.69 & 135.30$\pm$139.95 & 289.95$\pm$103.63\\
    ELTL &23.81$\pm$1.47 & 15.19$\pm$12.50 &4.12$\pm$0.11 & 299.14$\pm$202.21\\
    ECII &22.93$\pm$2.63 & 18.11$\pm$4.93 &6.45$\pm$1.42 & 37.94$\pm$28.25\\
    EvoLearner & 54.73$\pm$25.86 &48.00$\pm$31.38 &17.16$\pm$9.20 & 213.78$\pm$81.03\\
    \midrule
    NCES$_{LSTM}$ & 0.16$\pm$0.00 & 0.19$\pm$0.00 & 0.08$\pm$0.00 & 0.13$\pm$0.00\\
    NCES$_{GRU}$ & \underline{0.15$\pm$0.00} & \underline{0.18$\pm$0.00} & \underline{0.08$\pm$0.00} & \underline{0.06$\pm$0.00}\\
    NCES$_{ST}$ & \textbf{0.08$\pm$0.00}* & \textbf{0.11$\pm$0.00}* & \textbf{0.07$\pm$0.00}* & \textbf{0.04$\pm$0.00}*\\
    NCES$_{ST+GRU}$ & 0.16$\pm$0.00 & 0.25$\pm$0.00 & 0.11$\pm$0.00 & 0.09$\pm$0.00\\
    NCES$_{ST+LSTM}$ & 0.23$\pm$0.00 & 0.23$\pm$0.00 & 0.11$\pm$0.00 & 0.11$\pm$0.00\\
    NCES$_{GRU+LSTM}$ & 0.24$\pm$0.00 & 0.32$\pm$0.00 & 0.13$\pm$0.00 & 0.17$\pm$0.00\\
    NCES$_{ST+GRU+LSTM}$ & 0.27$\pm$0.00 & 0.31$\pm$0.00 & 0.15$\pm$0.00 & 0.15$\pm$0.00\\
    \bottomrule
  \end{tabular}
  \label{tab:evaluation-conex}
\end{table}
We compare our approach against EvoLearner, CELOE, ECII, ELTL. The maximum execution time for CELOE and EvoLearner was set to 300~seconds per learning problem while ECII and ELTL were executed with their default settings, as they do not have the maximum execution time parameter in their original implementation. 
From Table~\ref{tab:evaluation-conex}, we can observe that our approach (with ensemble prediction) significantly outperforms all other approaches in runtime on all datasets, and in F-measure on Carcinogenesis and Vicodi. 
\begin{table}[tb]
\centering
\caption{Evaluation results using TransE embeddings.}
\setlength{\tabcolsep}{1pt}
\footnotesize
  \begin{tabular}{@{}l@{\hskip 3pt}c@{\hskip 3pt}c@{\hskip 3pt}c@{\hskip 3pt}c@{}}
    \toprule
    & \multicolumn{4}{c}{\textbf{F$_1$ (\%) $\uparrow$}}\\
    \cmidrule(lr){2-5}
    & Carcinogenesis & Mutagenesis & Semantic Bible & Vicodi\\
    \midrule
    NCES$_{LSTM}$ & 84.52$\pm$27.01 & 76.45$\pm$34.22 & 59.63$\pm$34.98 & 79.06$\pm$30.07\\
    NCES$_{GRU}$ & 87.00$\pm$27.43 & 77.77$\pm$36.36 & 65.71$\pm$33.51 & 79.81$\pm$33.31\\
    NCES$_{ST}$ & 86.66$\pm$30.13 & 79.12$\pm$34.23 & 68.14$\pm$37.17 & 78.33$\pm$36.35\\
    NCES$_{ST+GRU}$ & 97.47$\pm$\phantom{0}9.63 & 91.05$\pm$21.11 & 75.36$\pm$32.88 & 89.27$\pm$25.09\\
    NCES$_{ST+LSTM}$ & 92.85$\pm$21.88 & 90.01$\pm$19.50 & 76.26$\pm$34.07 & 89.37$\pm$24.46\\
    NCES$_{GRU+LSTM}$ & 91.83$\pm$21.03 & 81.22$\pm$33.05 & 76.17$\pm$31.53 & 87.68$\pm$24.54\\
    NCES$_{ST+GRU+LSTM}$ & 97.56$\pm$11.55 & 91.24$\pm$21.27 & 85.82$\pm$24.85 & 90.05$\pm$23.67\\
    \bottomrule
    \toprule
    & \multicolumn{4}{c}{\textbf{Accuracy (\%)} $\uparrow$}\\
    \cmidrule(lr){2-5}
    & Carcinogenesis & Mutagenesis & Semantic Bible & Vicodi\\
    \midrule
    NCES$_{LSTM}$ & 97.40$\pm$14.19 & 96.59$\pm$11.27 & 84.10$\pm$24.91 & 97.17$\pm$11.98\\
    NCES$_{GRU}$ & 97.68$\pm$12.82 & 95.20$\pm$15.08 & 82.85$\pm$29.44 & 95.86$\pm$16.27\\
    NCES$_{ST}$ & 93.59$\pm$23.06 & 93.11$\pm$21.15 & 82.47$\pm$33.34 & 91.70$\pm$25.01\\
    NCES$_{ST+GRU}$ & 99.35$\pm$\phantom{0}5.52 & 99.07$\pm$\phantom{0}2.61 & 85.62$\pm$27.76 & 96.45$\pm$15.17\\
    NCES$_{ST+LSTM}$ & 98.11$\pm$11.18 & 99.00$\pm$\phantom{0}2.69 & 88.95$\pm$23.69 & 95.05$\pm$18.01\\
    NCES$_{GRU+LSTM}$ & 98.91$\pm$\phantom{0}9.46 & 96.08$\pm$13.64 & 89.25$\pm$24.54 & 97.37$\pm$12.60\\
    NCES$_{ST+GRU+LSTM}$ & 98.99$\pm$\phantom{0}9.45 & 98.90$\pm$\phantom{0}4.70 & 93.53$\pm$18.24 & 95.01$\pm$18.53\\
    \toprule
    & \multicolumn{4}{c}{\textbf{Runtime (sec.) $\downarrow$}}\\
    \cmidrule(lr){2-5}
    & Carcinogenesis & Mutagenesis & Semantic Bible & Vicodi\\
    \midrule
    NCES$_{LSTM}$ & 0.09$\pm$0.00 & 0.14$\pm$0.00 & 0.06$\pm$0.00 & 0.12$\pm$0.00\\
    NCES$_{GRU}$ & 0.05$\pm$0.00 & 0.15$\pm$0.00 & 0.06$\pm$0.00 & 0.13$\pm$0.00\\
    NCES$_{ST}$ & 0.04$\pm$0.00 & 0.09$\pm$0.00 & 0.05$\pm$0.00 & 0.05$\pm$0.00\\
    NCES$_{ST+GRU}$ & 0.08$\pm$0.00 & 0.18$\pm$0.00 & 0.08$\pm$0.00 & 0.11$\pm$0.00\\
    NCES$_{ST+LSTM}$ & 0.09$\pm$0.00 & 0.16$\pm$0.00 & 0.08$\pm$0.00 & 0.11$\pm$0.00\\
    NCES$_{GRU+LSTM}$ & 0.15$\pm$0.00 & 0.22$\pm$0.00 & 0.10$\pm$0.00 & 0.15$\pm$0.00\\
    NCES$_{ST+GRU+LSTM}$ & 0.14$\pm$0.00 & 0.22$\pm$0.00 & 0.11$\pm$0.00 & 0.14$\pm$0.00\\
    \bottomrule
  \end{tabular}
  \label{tab:evaluation-transe}
\end{table}
Table~\ref{tab:evaluation-transe} shows that NCES performs slightly better with ConEx embeddings than TransE embeddings except on the Carcinogenesis dataset. The standard deviation of NCES's prediction time is 0 because it performs batch predictions, i.e., it predicts solutions for all learning problems at the same time. The prediction time is averaged across learning problems and is therefore the same for each learning problem. We used the Wilcoxon Rank Sum test with a significance level of 5$\%$ and the null hypothesis that the compared quantities per dataset are from the same distribution. The best search-based approaches (CELOE and EvoLearner) only outperform NCES instances (including ensemble models) on the smallest datasets (Semantic Bible and Mutagenesis). The reason for this is that deep learning models are data-hungry and often fail to generalize well on small datasets. Our approach is hence well suited for large knowledge bases where search-based approaches are prohibitively slow. 
\subsection{Discussion}
\label{sec:discussion}
The hypothesis behind this work was that high-quality class expressions can be synthesized directly out of training data, i.e., without the need for an extensive search. Our results clearly undergird our hypothesis. While NCES is outperformed by CELOE and EvoLearner on small datasets, it achieves the best performance on Carcinogenesis with over 5\% absolute improvement in F-measure. This large difference is due to the fact that most search-based approaches fail to find any suitable solution for some learning problems. For example, the first learning problem on the Vicodi knowledge base is (\texttt{Disaster} $\sqcup$ \texttt{Military-Organisation}) $\sqcap$ ($\neg$\texttt{Engineer}). The solutions computed by each of the approaches are as follows: CELOE: \texttt{Flavour} $\sqcap$ ($\neg$\texttt{Battle}) $\sqcap$ ($\neg$\texttt{Person}) [F$_1$: 1.95\%], ELTL: \texttt{Flavour} $\sqcap$ ($\exists$ \texttt{related.($\exists$ related.Role)}) [F$_1$: 0\%], ECII: \texttt{Organisation} $\sqcup$ $\neg$\texttt{VicodiOI} [F$_1$: 13.16\%], EvoLearner: \texttt{Military-Organisation} $\sqcap$ \texttt{Military-Organisation} [F$_1$: 73.17\%], and NCES$_{ST+GRU}$\footnote{Here, NCES uses ConEx embeddings}: (\texttt{Disaster} $\sqcup$ \texttt{Military-Organisation}) $\sqcap$ ($\neg$\texttt{Engineer}) [F$_1$: 100\%]. Here, our ensemble model NCES$_{ST+GRU}$ synthesized the exact solution, which does not appear in the training data of NCES, while the best search-based approach, EvoLearner, could only compute an approximate solution with an F-measure of 73.17\%. On the other hand, CELOE, ECII, and ELTL failed to find any suitable solutions within the set timeout. 

The scalability of the synthesis step of our approaches makes them particularly suitable for situations where many class expressions are to be computed for the same knowledge base. For example, taking into account the average training and inference time of the Set Transformer architecture, one can conjecture that the minimum number of learning problems from which the cost of deep learning becomes worthwhile is: 11 for NCES vs. CELOE, 25 for NCES vs. EvoLearner, 24 for NCES vs. ELTL, and 96 for NCES vs. ECII. These values are calculated by solving for $n$ in $ n \times T_{\texttt{algo\_learn}} > T_{\texttt{train}}+T_{\texttt{inference}} $, where $T_{\texttt{algo\_learn}}$, $T_{\texttt{train}}$, and $T_{\texttt{inference}}$ are the average learning time of a search-based approach, the training time, and the inference time of NCES, respectively. 

\section{Conclusion and Future Work}
We propose a novel family of approaches for class expression learning, which we dub neural class expression synthesizers (NCES). NCES use neural networks to directly synthesize class expressions from input examples without requiring an expensive search over all possible class expressions. Given a set timeout per prediction, we showed that our approach outperforms all state-of-the-art search-based approaches on large knowledge bases. Taking training time into account, our approach is suitable for application scenarios where many concepts are to be learned for the same knowledge base. In future work, we will investigate means to transfer the knowledge acquired on one knowledge base to other knowledge bases. Furthermore, we plan to extend our approach to more expressive description logics such as $\mathcal{ALCHIQ(D)}$.
{
\raggedbottom
\raggedright
\bibliographystyle{splncs04nat}
\bibliography{nces}
}
\end{document}